\documentclass[runningheads]{llncs}
\usepackage{graphicx}

\usepackage{tikz}
\usepackage{comment}
\usepackage{amsmath,amssymb} 
\usepackage{color}
\usepackage{amsmath}
\usepackage{amssymb}
\usepackage[accsupp]{axessibility}  
\usepackage{multirow}
\usepackage{amsfonts}
\usepackage{graphicx}
\usepackage{amsmath}
\usepackage{booktabs}
\usepackage{bm}
\usepackage{algorithm}
\usepackage{algorithmic}

\newcommand{\ie}{\textit{i}.\textit{e}.}
\newcommand{\eg}{\textit{e}.\textit{g}.}
\usepackage[pagebackref,breaklinks,colorlinks]{hyperref}
\usepackage[misc]{ifsym}
\begin{document}
\sloppy
\pagestyle{headings}
\mainmatter
\def\ECCVSubNumber{6623}  

\title{Real Spike: Learning Real-valued Spikes for Spiking Neural Networks} 

\titlerunning{Real Spike}
%
\author{Yufei Guo\thanks{Equal contribution.} \and Liwen Zhang$^{\star}$ \and Yuanpei Chen \and Xinyi Tong \and Xiaode Liu \and YingLei Wang \and Xuhui Huang\textsuperscript{\Letter} \and Zhe Ma\textsuperscript{\Letter}}
\authorrunning{Guo, Y. et al.}
%
\institute{Intelligent Science \& Technology Academy of CASIC, Beijing 100854, China \\
\email{yfguo@pku.edu.cn, lwzhang9161@126.com, starhxh@126.com, mazhe\_thu@163.com}} 
\maketitle

\begin{abstract}
	Brain-inspired spiking neural networks (SNNs) have recently drawn more and more attention due to their event-driven and energy-efficient characteristics. The integration of storage and computation para-digm on neuromorphic hardwares makes SNNs much different from Deep Neural Networks (DNNs). In this paper, we argue that SNNs may not benefit from the weight-sharing mechanism, which can effectively reduce parameters and improve inference efficiency in DNNs, in some hardwares, and assume that an SNN with unshared convolution kernels could perform better. Motivated by this assumption, a training-inference decoupling method for SNNs named as \textbf{Real Spike} is proposed, which not only enjoys both unshared convolution kernels and binary spikes in inference-time but also maintains both shared convolution kernels and \textbf{\underline{Real}}-valued \textbf{\underline{Spike}}s during training. This decoupling mechanism of SNN is realized by a re-parameterization technique. Furthermore, based on the training-inference-decoupled idea, a series of different forms for implementing \textbf{Real Spike} on different levels are presented, which also enjoy shared convolutions in the inference and are friendly to both neuromorphic and non-neuromorphic hardware platforms. A theoretical proof is given to clarify that the Real Spike-based SNN network is superior to its vanilla counterpart. Experimental results show that all different \textbf{Real Spike} versions can consistently improve the SNN performance. Moreover, the proposed method outperforms the state-of-the-art models on both non-spiking static and neuromorphic datasets.
\end{abstract}

\section{Introduction}

\begin{figure}[t]
	\centering
	\includegraphics[width=0.48\textwidth]{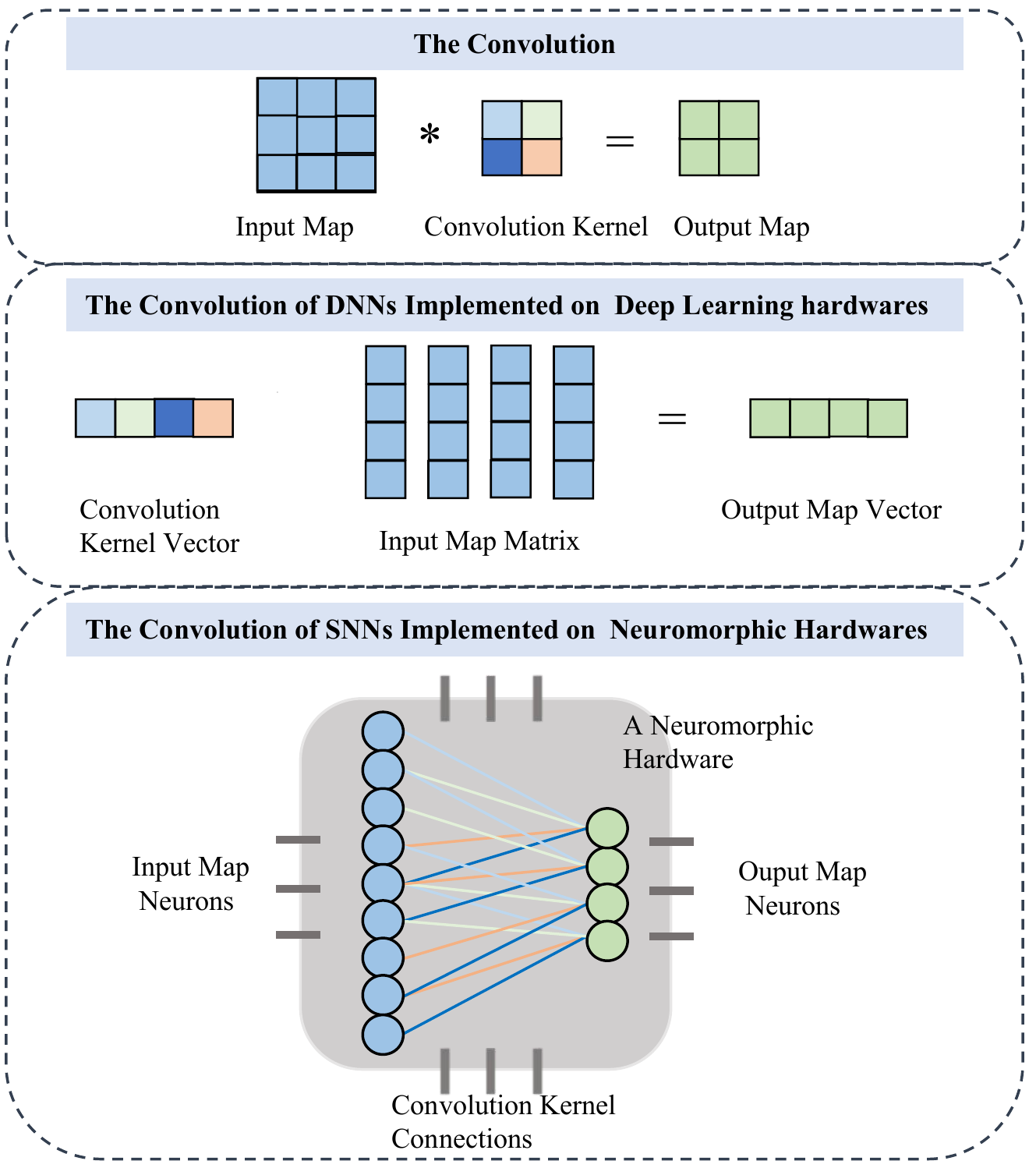} 
	\caption{The difference of convolution computing process between DNNs and SNNs. For DNNs, the calculation is conducted in a highly-paralleled way on conventional hardwares. However, for SNNs, each connection between neurons will be mapped into a synapse on some neuromorphic hardwares, which cannot benefit from the advantages of the weight-shared convolution kernel,~\eg, inference acceleration and parameter reduction.}
	\label{fig:difference}
\end{figure}

Spiking Neural Networks (SNNs) have received increasing attention as a novel brain-inspired computing model that adopts binary spike signals to communicate between units. Different from the Deep Neural Networks (DNNs), SNNs transmit information by spike events, and the computation dominated by the addition operation occurs only when the unit receives spike events. Benefitting from this characteristic, SNNs can greatly save energy and run efficiently when implementing on neuromorphic hardwares,~\eg, SpiNNaker~\cite{2008SpiNNaker}, TrueNorth~\cite{2015TrueNorth}, Darwin~\cite{2015Darwin}, Tianjic~\cite{2019Towards}, and Loihi~\cite{2018Loihi}.


The success of DNNs inspires the SNNs in many ways. Nonetheless, the rich spatio-temporal dynamics, event-driven paradigm, and friendly to neuromorphic hardwares make SNNs much different from DNNs, and directly applying the successful experience of DNNs to SNNs may limit the performance of SNNs. As one of the most widely used techniques in DNNs, the weight-shared convolution kernel shows great advantages. It can reduce the parameters of the network and accelerate the inference. However, SNNs show great advantages in the condition of being implemented on neuromorphic hardwares which is very different from DNNs being implemented on deep learning hardwares. As shown in Fig.~\ref{fig:difference}, in DNNs, the calculation is carried out in a highly-paralleled way on deep learning hardwares, thus sharing convolution kernel can improve the computing efficiency by reducing the data transferring between separated memory and processing units. However, for an ideal situation, to take full advantage of the storage-computation-integrated paradigm of neuromorphic hardwares, each unit and connection of the SNNs in the inference phase should be mapped into a neuron and synapse in neuromorphic hardware, respectively. Though these hardwares could be multiplexed, it also increases the complexity of deploymention and extra cost of data transfer. As far as we know, at least Darwin~\cite{2015Darwin}, Tianjic~\cite{2019Towards}, and other memristor-enabled neuromorphic computing systems~\cite{articleFully} adopt this one-to-one mapping form at present. Hence, all the components of an SNN will be deployed as a fixed configuration on these hardwares, no matter they share the same convolution kernel or not. Unlike the DNNs, the shared convolution kernels will not bring SNNs the advantages of parameter reduction and inference acceleration in this situation. Hence we argue that it would be better to learn unshared convolution kernels for each output feature map in SNNs.

Unfortunately, whether in theory or technology, it is not feasible to directly train an unshared convolution kernels-based SNN. First, there is no obvious proof that learning different convolution kernels directly will surely benefit the network performance. Second, due to the lack of mature development platforms for SNNs, many efforts are focusing on training SNNs with DNN-oriented programming frameworks, which usually do not support the learning of unshared convolution kernels for each feature map directly. Considering these limitations, we focus on training SNNs with unshared convolution kernels based on the modern DNN-oriented frameworks indirectly.

\begin{figure}[t]
	\centering
	\includegraphics[width=0.95\textwidth]{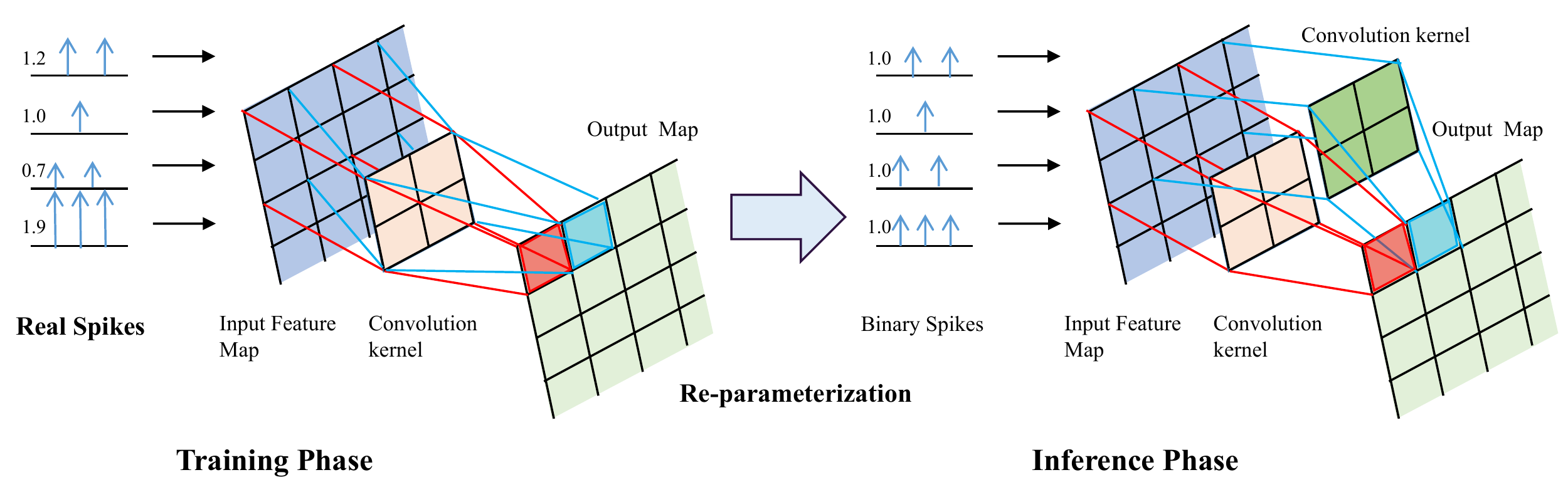} 
	\caption{The overall workflow of \textbf{Real Spike}. In the training phase, the SNN learns real-valued spikes and the shared convolution kernel for each output feature map. In the inference phase, the real spikes can be converted to binary spikes via convolution kernel re-parameterization. Then an SNN with different convolution kernels for every feature map is obtained.}
	\label{fig:workflow}
\end{figure}

Driven by the above reasons, a training-time and inference-time decoupled SNN is proposed, where a neuron can emit \emph{real-valued spikes} during training but binary spikes during inference, dubbed \textbf{Real Spike}. The training-time real-valued spikes can be converted to inference-time binary spikes via convolution kernel re-parameterization and a shared convolution kernel, which can be derived into multiples then (see details in  Sec. 3.3). In this way, an SNN with different convolution kernels for every output feature map can be obtained as we expected. Specifically, in the training phase, the SNN will learn real-valued spikes and a shared convolution kernel for every output feature map. While in the inference phase, every real-valued spike will be transformed into a binary spike by folding a part of the value to its corresponding kernel weight. Due to the diversity of the real-valued spikes, by absorbing part of the value from each real spike, the original convolution kernel shared by each output map can be converted into multiple forms. Thus different convolution kernels for each feature map of SNNs can be obtained indirectly. It can be guaranteed theoretically that the \textbf{Real Spike} method can improve the performance due to the richer representation capability of real-valued spikes than binary spikes (see details in  Sec. 3.4). Besides, \textbf{Real Spike} is well compatible with present DNN-oriented programming frameworks, and it still retains the advantages of DNN-oriented frameworks in terms of the convolution kernel sharing mechanism in the training. Furthermore, we extract the essential idea of training-inference-decoupled and extend \textbf{Real Spike} to a more generalized form, which is friendly to both neuromorphic and non-neuromorphic hardwares (see details in  Sec. 3.5). The overall workflow of the proposed method is illustrated in Fig. \ref{fig:workflow}.

Our main contributions are summarized as follows:
\begin{itemize}
	\item We propose the \textbf{Real Spike}, a simple yet effective method to obtain SNNs with unshared convolution kernels. The Real Spike-SNN can be trained in DNN-oriented frameworks directly. It can effectively enhance the information representation capability of the SNN without introducing training difficulty.
	
	\item The convolution kernel re-parameterization is introduced to decouple a training-time SNN with real-valued spikes and shared convolution kernels, and an inference-time SNN with binary spikes and unshared convolution kernels.
		
	\item We extend the \textbf{Real Spike} to other different granularities (layer-wise and channel-wise). These extensions can keep shared convolution kernels in the inference and show advantages independent of specific hardwares.
	
	\item The effectiveness of \textbf{Real Spike} is verified on both static and neuromorphic datasets. Extensive experimental results show that our method performs remarkably.
\end{itemize}

\section{Related Work}

This work starts from training more accurate SNNs with unshared convolution kernels for each feature map. Considering the lack of specialized suitable platforms that can support the training of deep SNNs, powerful DNN-oriented programming frameworks are adopted. To this end, we briefly overview recent works of SNNs in three aspects: (i) learning algorithms of SNNs; (ii) SNN programming frameworks; (iii) convolutions.

\subsection{Learning Algorithms of SNNs}

The learning algorithms of SNNs can be divided into three categories: converting ANN to SNN (ANN2SNN)~\cite{2015Spiking,2020Deep,2019Going,li2021free}, unsupervised learning~\cite{Peter2015Unsupervised,2018ABiologically}, and supervised learning~\cite{2017articleGradient,2019Surrogate,li2021differentiable,Shrestha2018,Guo_2022_CVPR}. ANN2SNN converts a pre-trained ANN to an SNN by transforming the real-valued output of the activation function to binary spikes. Due to the success of ANNs, ANN2SNN can generate an SNN in a short time with competitive performance. However, the converted ANN inherits the limitation of ignoring rich temporal dynamic behaviors from DNNs, it cannot handle neuromorphic datasets well. On the other hand, ANN2SNN usually requires hundreds of timesteps to approach the accuracy of pre-trained DNNs. Unsupervised learning is considered a more biologically plausible method. Unfortunately, the lack of sufficient understanding of the biological mechanism prevents the network from going deep, thus it is usually limited to small datasets and non-ideal performance. Supervised learning trains SNNs with error backpropagation. Supervised learning-based methods can not only achieve high accuracy within very few timesteps but also handle neuromorphic datasets. Our work falls under this category.

\subsection{SNN Programming Frameworks}

There exist several specialized programming frameworks for SNN modeling. However, most of them cannot support the direct training of deep SNNs. NEURON~\cite{Hines2006The}, a simulation environment for modeling computational models of neurons and networks of neurons, mainly focuses on simulating neuron issues with complex anatomical and physiological characteristics, which is more suitable for neuroscience research. BRIAN2~\cite{2009The} and NEST~\cite{2007NEST}, simulators for SNNs, aim at making the writing of simulation code as quick and easy as possible for the user, but they are not designed for the supervised learning for deep SNNs. SpikingJelly~\cite{SpikingJelly}, an open-source deep learning framework for SNNs based on PyTorch, provides a solution to establish SNNs by mature DNN-oriented programming frameworks. However, so far most functions of the SpikingJelly are still under development. Instead of developing a new framework, we attempt to investigate an ingenious solution to develop the SNNs that have unshared convolutional kernels for each output feature map in the DNN-oriented frameworks, which have demonstrated an easy-to-use interface.

\subsection{Convolutions}

Convolutional Neural Networks have recently harvested a huge success in large-scale computer vision tasks~\cite{1998Gradient,2012ImageNetNIPS,2014Very,2014Going,2016Deep,2017Faster}. Due to the abilities of input scale adaptation, translation invariance, and parameter reduction, the stacked convolution layers help train a deep and well-performed neural network with fewer parameters compared to dense-connected fully connected layers. For a convolution layer, feature maps of the previous layer are convolved with some learnable kernels and presented through the activation function to form the output feature maps as the current layer. Each learnable kernel is corresponding to an output map. In general, we have that
\begin{equation}
	\textbf{X}_{i,j} = f(\textbf{I}_j*\textbf{K}_i),
	\label{eq:conv}
\end{equation}
where $\textbf{I}_j \in \mathbb{R}^{k\times k\times C}$ denotes $j$-th block of the input maps with total $C$ channels, $\textbf{K}_i \in\mathbb{R}^{k\times k \times C}$ is the $i$-th convolution kernel with $i=1,~\dots,~C$, then $\textbf{X}_{i,j} \in \mathbb{R}$ is the $j$-th element in $i$-th output feature map. It can be seen that all the outputs in each channel share a same convolution kernel. However, when implementing a convolution-based SNN on above mentioned neuromorphic hardwares, all the kernels will be mapped as synapses no matter they are shared or not. Hence, keeping shared convolution kernel cannot show the same advantages for SNNs as DNNs. We argue that learning different convolution kernels for each output map may improve the performence of the SNN further. In this case, a convolution layer for SNNs can be written as
\begin{equation}\label{eq1}
	\textbf{X}_{i,j} = f(\textbf{I}_j*\textbf{K}_{i,j}),
\end{equation}
where $\textbf{K}_{i,j} \in\mathbb{R}^{k\times k \times C}$ is the $j$-th convolution kernel for $i$-th output map.
Unfortunately, it is not easy to directly implement the learning of unshared convolution kernel for SNNs in the DNN-oriented frameworks. And dealing with this issue is one of the important works in this paper. 

\section{Materials and Methodology}

Aiming at training more accurate SNNs, we adopt the explicitly iterative leaky Integrate-and-Fire (LIF) model, which can be implemented on mature DNN-oriented frameworks easily to simulate the fundamental computing unit of SNNs. Considering that it is not easy to realize unshared convolution kernels for each output feature map with existing DNN-oriented frameworks, a modified LIF model that can emit the real-valued spike is proposed first. By using this modified LIF model, our SNNs will learn real spikes along with the shared convolution kernel for every output channel as DNNs during training. While for the inference phase, real spikes will be transformed into binary spikes and each convolution kernel will be re-parameterized as multiple different convolution kernels, so the advantages of SNNs can be recovered and unshared convolution kernel for each output map can be obtained. Then, to make the proposed design more general, we also propose layer-wise and channel-wise \textbf{Real Spike}, which can keep shared convolution kernels in both training phase and inference phase and will introduce no more parameters than its vanilla counterpart. In this section, we will introduce explicitly iterative LIF model, \textbf{Real Spike}, re-parameterization, and extensions of \textbf{Real Spike} successively.

\subsection{Explicitly Iterative Leaky Integrate-and-Fire Model}

The spiking neuron is the fundamental computing unit of SNNs. The LIF neuron is most commonly used in supervised learning-based methods. Mathematically, a LIF neuron can be described as 
\begin{equation}\label{eq1}
	\tau \frac{\partial u}{\partial t} = -u+I ,\quad u < V_{th}
\end{equation}
\begin{equation}\label{eq2}
	u=u_{rest} \quad \& \quad fire \; a \; spike,\quad u \ge V_{th}
\end{equation}
where $u$, $u_{rest}$, $\tau$, $I$, and $V_{th}$ represent the membrane potential, membrane resting potential, membrane time constant that controls the decaying rate of $u$, pre-synaptic input, and the given firing threshold, respectively. When $u$ is below $V_{th}$, it acts as a leaky integrator of $I$. On the contrary, when $u$ exceeds $V_{th}$, it will fire a spike and propagate the spike to the next layer, then it will be reset to the resting potential, $u_{rest}$, which is usually set as $0$.

The LIF model has a complex dynamics structure that is incompatible with nowadays DNN-oriented framework. By discretizing and transforming the LIF model to an explicitly iterative LIF model, SNNs can be implemented in these mature frameworks. The hardware friendly iterative LIF model can be described as
\begin{equation}
	u(t) = \tau u(t-1)+I(t), \quad u(t) < V_{th}
	\label{eq:modlif}
\end{equation}
\begin{equation}
	o(t)=
	\left\{
	\begin{array}{lll}
		1, \ \ if \; u(t) \ge V_{th}, \\
		0, \ \ otherwise.
	\end{array}
	\right.
	\label{eq:firespike}
\end{equation}
where $V_{th}$ is set to 0.5 in this work. Up to now, there is still an obstacle for training SNNs in a direct way,~\ie, Eq.~\eqref{eq:firespike} is non-differentiable. As in other work~\cite{2020DIET,2018Direct,2020LISNN}, we appoint a rectangular function as the particular pseudo derivative of spike firing as follows,
\begin{equation}\label{ste}
	\frac{do}{du}=
	\left\{
	\begin{array}{lll}
		1, \ \ if \; 0 \le u \le 1, \\
		0, \ \ otherwise.
	\end{array}
	\right.
\end{equation}

With all these settings, now we can train an SNN on DNN-oriented frameworks.

\subsection{Real Spike}

As aforementioned, driven by the suppressed advantages of the shared convolution kernel and the expectation of enhancing the capacity of information representation for SNNs, we turn the problem of learning unshared convolution kernels to learning real spikes. To be more specific, the output of our modified LIF model in Eq.~\eqref{eq:firespike} is further rewritten as
\begin{equation}
	\tilde{o}(t)=a \cdotp o(t)
	\label{eq:realspike}
\end{equation}
where $a$ is a learnable coefficient. With this modification, our LIF model can emit a \emph{real-valued spike}, dubbed \textbf{Real Spike}. Then we train SNNs with this modified LIF model along with the shared convolution kernel for each output map, which can be easily implemented in DNN-oriented frameworks. Obviously, unlike the binary spike, the real-valued spike will lose the advantage of computation efficiency of SNNs, since the corresponding multiplication cannot be substituted to addition. And another problem is that the learned convolution kernels for each output map are still shared at this time. Therefore, to jointly deal with these problems, we propose a training-inference decoupled framework, which can transform real spikes into binary spikes and convert the shared convolution kernel as different kernels by using re-parameterization, which will be introduced in the next subsection.

\begin{figure}[t]
	\centering	
	\includegraphics[width=0.8\textwidth]{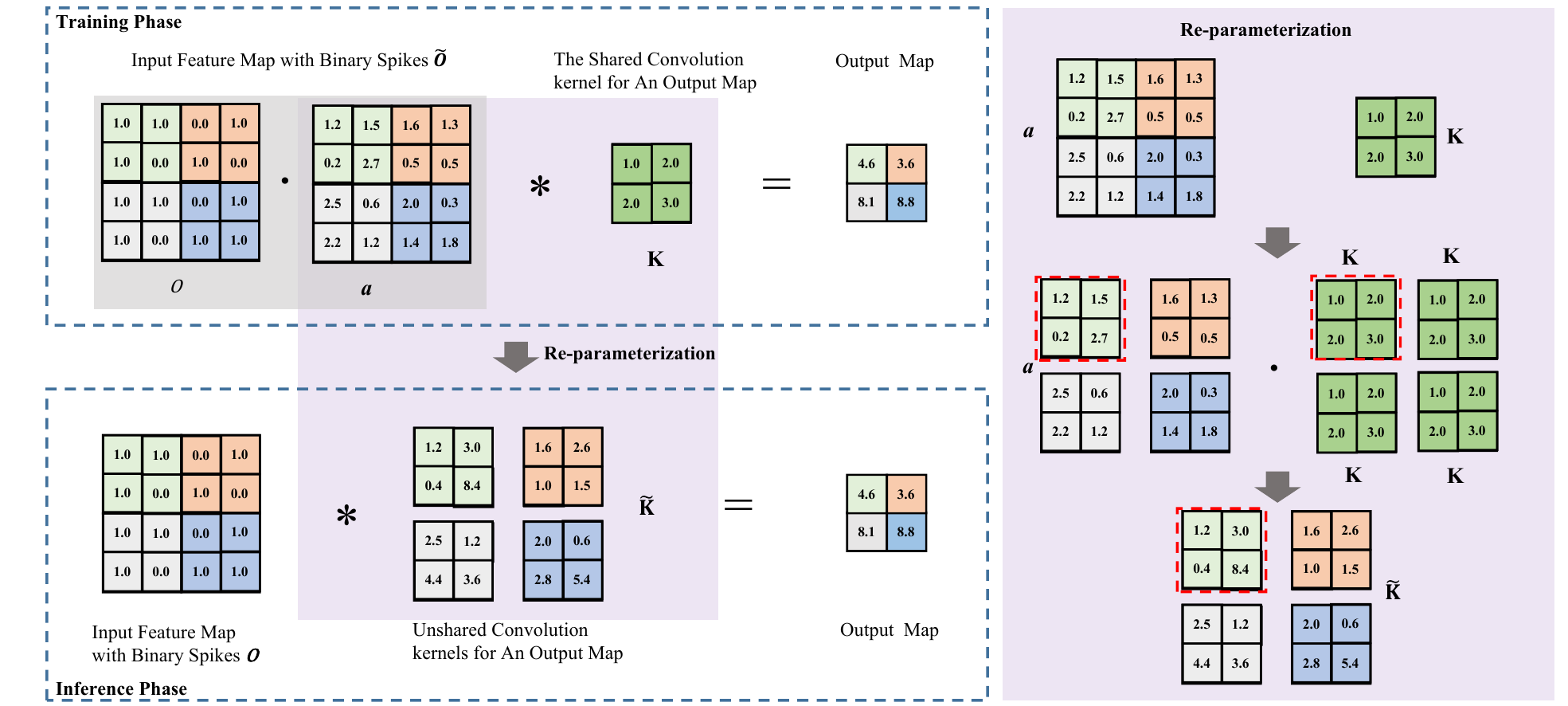} 
	\caption{The diagram of re-parameterization by a simple example.}
	\label{fig:Reparameterization}
\end{figure}

\subsection{Re-parameterization}\label{sec:param}

Consider a convolution layer, which takes $\textbf{F}\in\mathbb{R}^{D_F \times D_F \times M}$ as input feature maps, and generates the output feature maps, $\textbf{G}\in \mathbb{R}^{D_G \times D_G \times N}$, where $D_F$ and $M$ denote the size of the square feature maps and the number of channels (depths) for the input, respectively; $D_G$ and $N$ denote the size of the square feature maps and the number of channels (depths) for the output, respectively. The convolution layer is actually parameterized by a group of convolution kernels, which can be denoted as a tensor, $\textbf{K} \in\mathbb{R}^{D_K \times D_K \times M \times N}$ with a spatial dimension of $D_K$. Then, each element in $\textbf{G}$ is computed as
\begin{equation}
	\textbf{G}_{k,l,n}=\sum_{i,j,m} \textbf{K}_{i,j,m,n} \cdotp \textbf{F}_{k+i-1,l+j-1,m}
	\label{eq:output1}
\end{equation}
For standard SNNs, the elements of input maps are binary spikes, while in this work, the SNN is trained with real-valued spikes for the purpose of enhancing the network representation capacity. In this case, we can further denote the input feature map, $\textbf{F}$, according to~Eq.~\eqref{eq:realspike} as follows
\begin{equation}
	\textbf{F}=\textbf{a} \odot \textbf{B}
	\label{eq:decF}
\end{equation}
where $\textbf{B}$ and $\textbf{a}$ denote a binary tensor and a learnable coefficient tensor, respectively. With this element-wise multiplication in Eq.~\eqref{eq:decF}, we can extract a part of the value from each element in $\textbf{F}$, and fold it into the shared convolution kernel one-by-one according to the corresponding position during inference. Then the single shared convolution kernel can be turned into multiples without changing the values of the output maps. Through this decoupling process, a new SNN that can emit binary spikes and enjoy different convolution kernels will be obtained. This process can be illustrated from Eq.~\eqref{eq:k1} to Eq.~\eqref{eq:k3} as follows:
\begin{equation}\label{eq:k1}
	\textbf{G}_{k,l,n}=\sum_{i,j,m} \textbf{K}_{i,j,m,n} \cdotp (\textbf{a}_{k+i-1,l+j-1,m} \cdotp \textbf{B}_{k+i-1,l+j-1,m})
\end{equation}
\begin{equation}\label{eq:k2}
	\textbf{G}_{k,l,n}=\sum_{i,j,m} (\textbf{a}_{k+i-1,l+j-1,m} \cdotp \textbf{K}_{i,j,m,n}) \cdotp \textbf{B}_{k+i-1,l+j-1,m}
\end{equation}
\begin{equation}\label{eq:k3}
	\textbf{G}_{k,l,n}=\sum_{i,j,m} \tilde{\textbf{K}}_{k,l,i,j,m,n} \cdotp \textbf{B}_{k+i-1,l+j-1,m}
\end{equation}
where $\tilde{\textbf{K}}$ is the unshared convolution kernel tensor.

The whole process described above is called re-parameterization, which allows us to convert a real-valued-spike-based SNN into an output-invariant binary-spike-based SNN with unshared convolution kernels for each output map. That is, re-parameterization provides a solution to obtain an SNN with unshared convolution kernels under DNN-oriented frameworks by decoupling the training-time SNN and inference-time SNN. Figure \ref{fig:Reparameterization} illustrates the details of re-parameterization by a simple example.

\subsection{Analysis and Discussions}\label{sec:analysis}

In this work, we assume that firing real-valued spikes during training can help increase the representation capacity of the SNNs. To verify our assumption, a series of analyses and discussions are conducted by using the information entropy concept. Given a scalar, vector, matrix, or tensor, $\textbf{X}$, its representation capability is denoted as $\mathcal{R}(\textbf{X})$, which can be measured by the information entropy of $\textbf{X}$, as follows
\begin{equation}\label{eq:entro}
	\mathcal{R}(\textbf{X}) = \max \mathcal{H}(\textbf{X}) = \max (-\sum_{x \in \textbf X } p_{\textbf X}(x)logp_{\textbf X}(x))
\end{equation}
where $p_{\textbf X}(x)$ is the probability of a sample from $\textbf X$. When $p_{\textbf X}(x_1)= \cdots = p_{\textbf X}(x_n)$, $\mathcal{H}(\textbf{X})$ reaches its maximum (see Appendix A.1 for detailed proofs). For a binary spike $o$, it can be expressed with 1 bit, and the number of samples from $o$ is $2$. While the real-valued spike $\tilde{o}$ needs 32 bits, which consists of $2^{32}$ samples. Hence, $\mathcal{R}(o) = 1$ and $\mathcal{R}$ $(\tilde{o}) = 32$ according to Eq.~\eqref{eq:entro}. Obviously, the representation capability of real spikes far exceeds that of binary spikes. This indicates that real spikes will enhance the information expressiveness of SNNs, which accordingly benefit the performance improvement. To further show the difference between real spikes and binary spikes intuitively, the visualizations of some channels expressed by real spikes and binary spikes are given respectively in the appendix.

Another intuitive conjecture to explain why the SNN with real-valued spikes performs better than its counterpart with binary spikes is that, for the former one, the information loss can be restrained to some extent by changing the fixed spike value to an appropriate value with a scalable coefficient, $a$; while for the later one, the firing function would inevitably induce the quantization error.

\begin{figure}[t]
	\centering
	\includegraphics[width=0.6\textwidth]{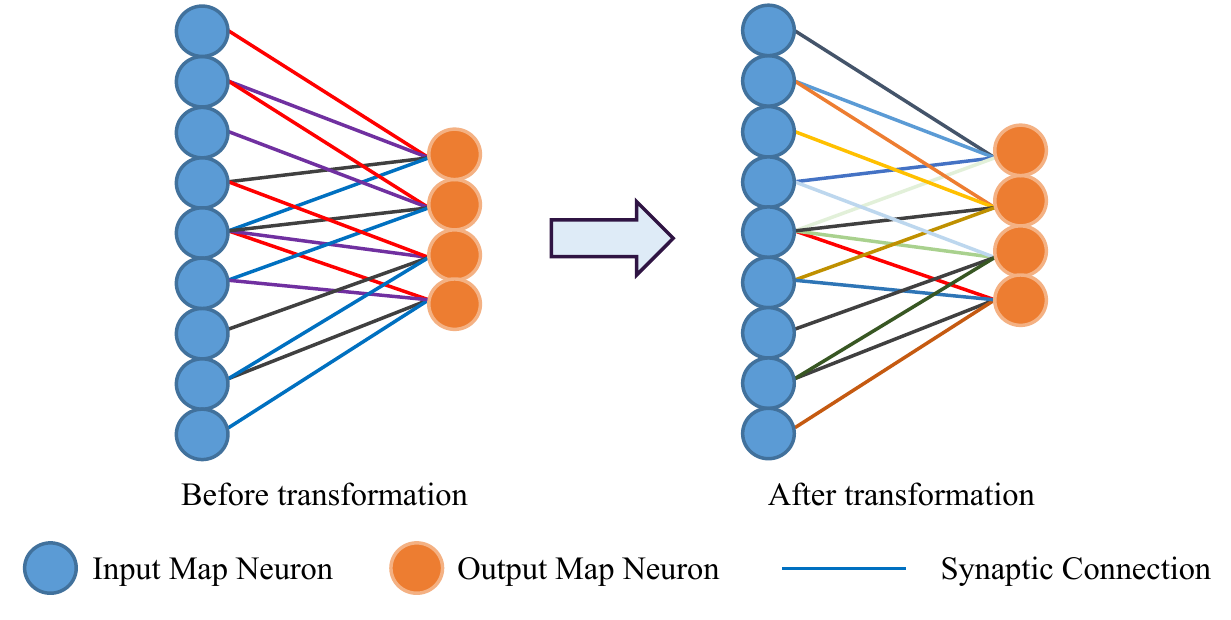} 
	\caption{The difference of adjacent layers in SNNs implemented on neuromorphic hardwares before re-parameterization and after re-parameterization. The convolution kernel re-parameterization will only change the values of the connections between neurons without introducing any additional computation and storage resource since the entire architecture topology of an SNN must be completely mapped into the hardwares. }
	\label{fig:mapping}
\end{figure}

It can be concluded from the above analysis that, learning real-valued spikes instead of the binary spikes in the training phase, enables us to train a more accurate SNN. And by performing re-parameterization in the inference phase, real spikes can be converted to binary spikes and unshared convolution kernels can be obtained. In another word, learning \textbf{Real Spike} is actually used to generate a better information encoder, while the re-parameterization will transfer the rich information encoding capacity from \textbf{Real Spike} into information decoding. Moreover, the transformed model after re-parameterization can also retain the advantage coming from the binary spike information processing paradigm in standard SNNs. As shown in Fig.~\ref{fig:mapping}, when deploying an SNN into some neuromorphic hardwares, all the units and their connections of the network will be mapped as neurons and synaptic connections one by one. Hence, the SNN with multiple different convolution kernels will not introduce any computation and storage resource.

\subsection{Extensions of Real Spike}

The key observation made in the \textbf{Real Spike} is that re-scaling the binary spike of the LIF neuron with a real-valued coefficient, $a$, can increase the representation capability and accurancy of SNNs. As shown in Eq.~\eqref{eq:decF}, the default Real Spike is performed in the element-wise way. In a similar fashion, we argue that introducing scaling coefficent by layer-wise or channel-wise manners will also retain the benefit to some extent. Then, we further propose to re-formulate Eq.~\eqref{eq:realspike} for one layer as follows
\begin{equation}\label{spikelayer}
	\tilde {\textbf{o}}(t)=\textbf{a} \cdotp \textbf{o}(t)
\end{equation}
With this new formulation we can explore various ways of introducing $\textbf{a}$ during training. Specifically, we propose to introduce $\textbf{a}$ for each layer in the following 3 ways:

\noindent \textbf{Layer-wise:}
\begin{equation}\label{hello}
	\textbf{a} \in \mathbb{R}^{1 \times 1 \times 1}
\end{equation}
\noindent \textbf{Channel-wise:}
\begin{equation}\label{hello}
	\textbf{a} \in \mathbb{R}^{C \times 1 \times 1}
\end{equation}
\noindent \textbf{Element-wise:}
\begin{equation}\label{hello}
	\textbf{a} \in \mathbb{R}^{C \times H \times H}
\end{equation}
where $C$ is the number of channels and $H$ is the spatial dimension of a square feature map. Obviously, for layer-wise and channel-wise Real Spike, re-parameterization will only re-scale the shared convolution kernels without transferring them to different ones. In these two forms, our \textbf{Real Spike} act in the same way as conventional SNNs on any hardwares. That is to say, starting from
the motivation of how to take full advantage of the integration of memory and computation, we propose element-wise \textbf{Real Spike}. Then we extract the essential idea of training-inference-decoupled and extend \textbf{Real Spike} to a more generalized form, which is friendly to both neuromorphic and non-neuromorphic hardware platforms.

\section{Experiment}

The performance of the \textbf{Real Spike}-based SNNs were evaluated on several traditional static datasets (CIFAR-10~\cite{CIFAR-10}, CIFAR-100~\cite{CIFAR-10}, ImageNet~\cite{2012ImageNetNIPS}) and one neuromorphic dataset (CIFAR10-DVS~\cite{2017CIFAR10}). And multiple widely-used spiking archetectures including ResNet20~\cite{2020DIET,2019Going}, VGG16~~\cite{2020DIET}, ResNet18~\cite{2021Deep}, and ResNet34~\cite{2021Deep} were used to verify the effectiveness of our \textbf{Real Spike}. Detailed introduction of the datasets and experimental settings are provided in the appendix. Extensive ablation studies were conducted first to compare the SNNs with real-valued spikes and their binary spike counterpart. Then, we comprehensively compared our SNNs with the existing state-of-the-art SNN methods.

\subsection{Ablation Study for Real Spike}

\begin{table}[t]
	\centering	
	\caption{Ablation study for \textbf{Real Spike}.}	
	\label{tab:ab}	
	\setlength{\tabcolsep}{6mm}{
	\begin{tabular}{llcc}	
		\toprule
		Dataset & Architecture & Timestep & Accuracy \\	
		\toprule
		\multirow{6}{*}{CIFAR10}	
		& \multirow{3}{*}{ResNet20 w/ BS} & 2 & 88.91\%   \\	
		&                                           & 4 & 91.73\%   \\
		&                                           & 6 & 92.98\%   \\
		\cline{2-4}
		& \multirow{3}{*}{ResNet20 w/ RS} & 2 & 90.47\%   \\	
		&                                           & 4 & 92.53\%   \\
		&                                           & 6 & 93.44\%   \\
		\hline
		\multirow{6}{*}{CIFAR100}	
		& \multirow{3}{*}{ResNet20 w/ BS} & 2 & 62.59\%   \\	
		&                                           & 4 & 63.27\%   \\
		&                                           & 6 & 67.12\%   \\
		\cline{2-4}
		& \multirow{3}{*}{ResNet20 w/ RS} & 2 & 63.40\%   \\	
		&                                           & 4 & 64.87\%   \\
		&                                           & 6 & 68.60\%   \\
		\bottomrule			   		         	            			         	
	\end{tabular}
	}
	\footnotesize
	\leftline{RS represents real spikes and BS represents binary spikes.}
\end{table}

\begin{figure}[t]
	\centering
	\includegraphics[width=0.346\textwidth]{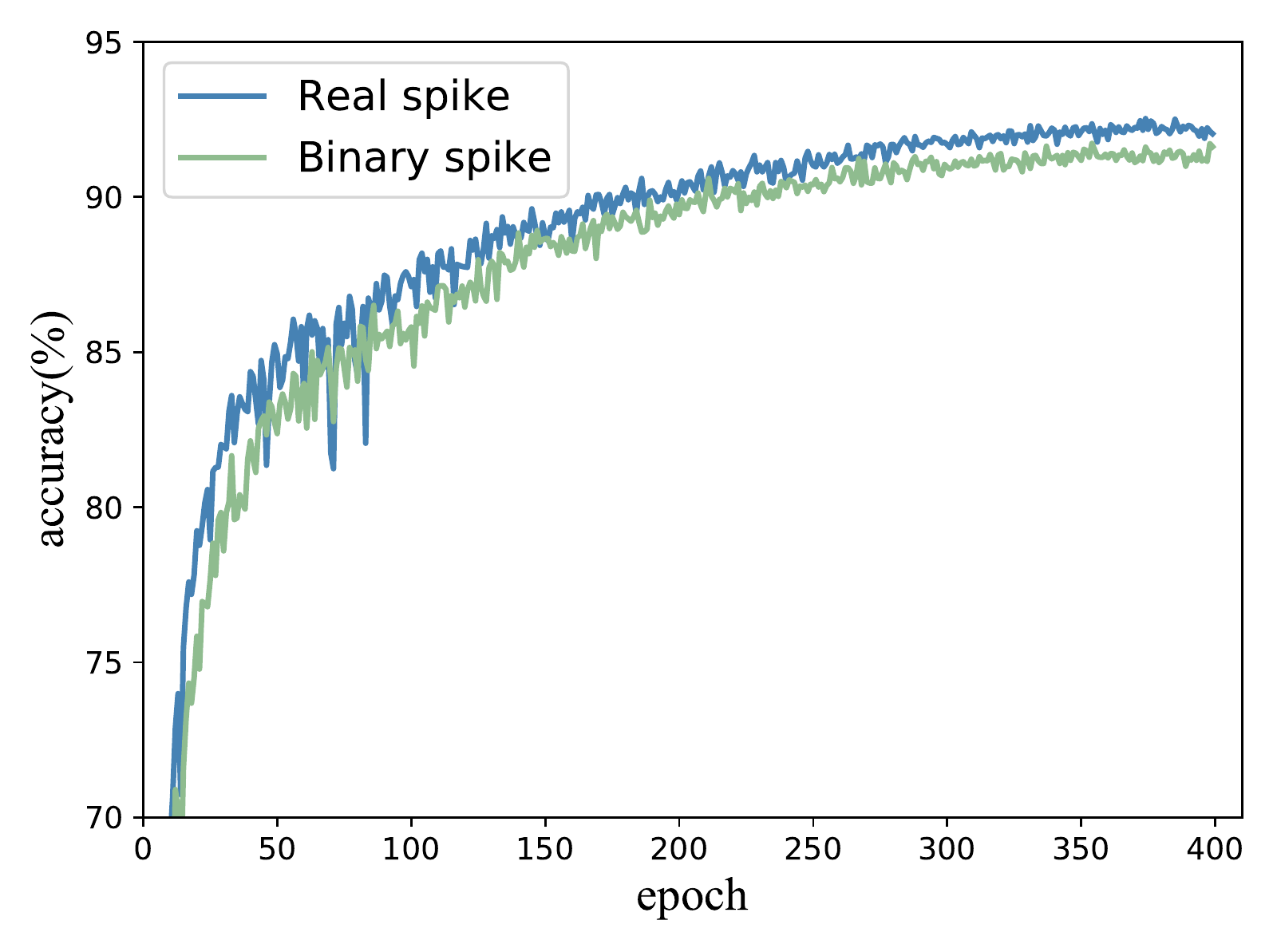}
	\includegraphics[width=0.346\textwidth]{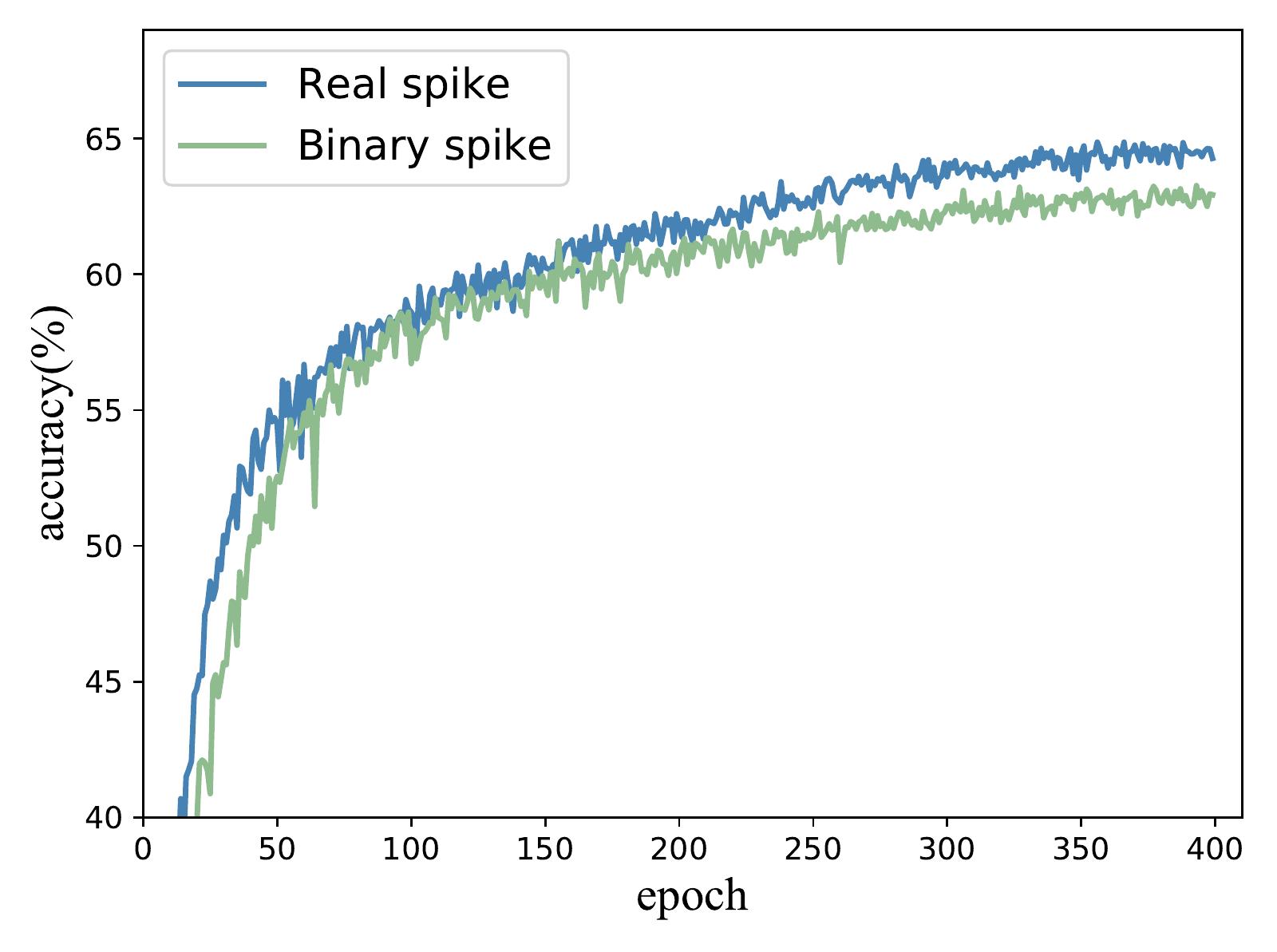}
	\caption{ The accuracy curves of ResNet20 with real-valued spikes and binary spikes with a timestep~=~4 on CIFAR-10(left) and CIFAR-100(right). The real-valued spikes-based SNNs obviously enjoy higher accuracy and easier convergence.}
	\label{fig:acc}
\end{figure}

The ablation study of \textbf{Real Spike} was conducted on CIFAR-10/100 datasets based on ResNet20. The SNNs with real-valued spikes and binary spikes were trained with the same configuration, with timestep varying from 2 to 6. Results in Tab.~\ref{tab:ab} show that the test accuracy of the SNNs with real-valued spikes is always higher than that with binary spikes. This is due to the richer representation capacity from Real Spike, which benefits the performance improvement of SNNs. Figure \ref{fig:acc} illustrates the test accuracy curves of ResNet20 with real-valued spikes and its counterpart with binary spikes on CIFAR-10/100 during training. It can be seen that the SNNs with real-valued spikes obviously perform better on the accuracy and convergence speed.

\subsection{Ablation Study for Extensions of Real Spike}

The performance of SNNs with different extensions of Real Spike was evaluated on CIFAR-10(100). Results in Tab.~\ref{tab:extension} show that element-wise \textbf{Real Spike} always outperforms the other versions. But all the Real Spike-based SNNs conformably outperform the vanilla (standard) SNN. Another observation is that the accuracy difference between layer-wise \textbf{Real Spike} and the binary spike is greater than that between layer-wise \textbf{Real Spike} and element-wise \textbf{Real Spike}. This shows that our method touches on the essence of improving the SNN accuracy and is very effective. 

\begin{table}[t]
	\centering	
	\caption{Performance comparison for different \textbf{Real Spike} versions.}	
	\label{tab:extension}
	\setlength{\tabcolsep}{6mm}{
	\begin{tabular}{llcc}	
		\toprule
		Dataset & Architecture & version & Accuracy \\	
		\hline
		\multirow{4}{*}{CIFAR10}	
		&\multirow{4}{*}{ResNet20} & Vanilla & 91.73\%   \\	
		&                                    & Layer-wise & 92.12\%   \\
		&                                    & Channel-wise & 92.25\%   \\	
		&                                    & Element-wise & 92.53\%   \\
		\hline
		\multirow{4}{*}{CIFAR100}	
		& \multirow{4}{*}{ResNet20} & Vanilla & 63.27\%   \\	
		&                                    & Layer-wise & 64.28\%   \\
		&                                    & Channel-wise & 64.71\%   \\	
		&                                    & Element-wise & 64.87\%   \\
		\bottomrule			   		         	            			         	
	\end{tabular}
	}
\end{table}

\subsection{Comparison with the State-of-the-art}

\begin{table}[tp]\small
	\centering	
	\caption{Performance comparison with SOTA methods.}	
	\label{tab:Comparison}	
	\begin{tabular}{llllcc}	
		\toprule
		Dataset & Method & Type & Architecture & Timestep & Accuracy \\	
		\toprule
		\multirow{20}{*}{CIFAR-10}	
		& SpikeNorm~\cite{2019Going} & ANN2SNN & VGG16 & 2500 & 91.55\%   \\	
		& Hybrid-Train~\cite{2020Enabling} & Hybrid  & VGG16 & 200 & 92.02\%   \\
		& SBBP~\cite{2019Enabling} & SNN training & ResNet11 & 100 & 90.95\%   \\
		& STBP~\cite{2018Direct} & SNN training & CIFARNet & 12 & 90.53\%   \\  	
		& TSSL-BP~\cite{2020Temporal} & SNN training & CIFARNet & 5 & 91.41\%   \\ 
		& PLIF~\cite{2020Incorporating} & SNN training & PLIFNet & 8 & 93.50\%   \\
		\cline{2-6}
		& \multirow{4}{*}{Diet-SNN~\cite{2020DIET}} & \multirow{4}{*}{SNN training} & \multirow{2}{*}{VGG16} & 5 & 92.70\%   \\ 
		& &  &  & 10 & 93.44\%   \\ 
		\cline{4-6}
		&  &  & \multirow{2}{*}{ResNet20} & 5 & 91.78\%   \\ 
		&  &  &  & 10 & 92.54\%   \\   
		\cline{2-6}
		& \multirow{3}{*}{STBP-tdBN~\cite{2020Going}} & \multirow{3}{*}{SNN training} & \multirow{3}{*}{ResNet19} 
		& 2 & 92.34\%   \\
		&  &  &											                                  & 4 & 92.92\%   \\
		&  &  &											                                   & 6 & 93.16\%   \\
		\cline{2-6}
		& \multirow{7}{*}{\textbf{Real Spike}} & \multirow{7}{*}{SNN training} & \multirow{3}{*}{ResNet19} 
		& 2 & \textbf{94.01\%}$\pm 0.10$  \\
		&  &  &											                                  & 4 & \textbf{95.60\%}$\pm 0.08$  \\
		&  &  &											                                   & 6 & \textbf{95.71\%}$\pm 0.07$   \\	
		\cline{4-6}	
		&  &          & \multirow{2}{*}{ResNet20} 
		& 5 & \textbf{93.01\%}$\pm 0.07$   \\
		&  &  &											                                  & 10 & \textbf{93.65\%}$\pm 0.05$   \\
		\cline{4-6}		
		&  &  &											     \multirow{2}{*}{VGG16} & 5 & \textbf{92.90\%}$\pm 0.09$  \\
		&  &  &											                              & 10 & \textbf{93.58\%}$\pm 0.06$   \\		
		\hline

		\multirow{10}{*}{CIFAR-100}	
		& BinarySNN~\cite{2020Exploring} & ANN2SNN & VGG15 & 62 & 63.20\%   \\
		& Hybrid-Train~\cite{2020Enabling} & Hybrid  & VGG11 & 125 & 67.90\%   \\  	
		& T2FSNN~\cite{2020T2FSNN} & ANN2SNN & VGG16 & 680 & 68.80\%   \\ 
		& Burst-coding~\cite{park2019fast} & ANN2SNN & VGG16 & 3100 & 68.77\%   \\
		& Phase-coding~\cite{Kim2018Deep} & ANN2SNN & VGG16 & 8950 & 68.60\%   \\
		\cline{2-6} 
		& \multirow{2}{*}{Diet-SNN~\cite{2020DIET}} & \multirow{2}{*}{SNN training} & ResNet20 & 5 & 64.07\%   \\ 
		&                                          &                                & VGG16 & 5 & 69.67\%   \\  
		\cline{2-6}
		& \multirow{3}{*}{\textbf{Real Spike} } & \multirow{3}{*}{SNN training} & {ResNet20} 
		& 5 & \textbf{66.60\%}$\pm 0.11$   \\
		&  &  &											                             {VGG16} & 5 & \textbf{70.62\%}$\pm 0.08$   \\
		&  &  &											                             {VGG16} & 10 & \textbf{71.17\%}$\pm 0.07$   \\
		\hline		
		\multirow{7}{*}{ImageNet}	
		& Hybrid-Train~\cite{2020Enabling} & Hybrid  & ResNet34 & 250 & 61.48\%   \\  
		& SpikeNorm~\cite{2019Going} & ANN2SNN & ResNet34 & 2500 & 69.96\%   \\		
		& STBP-tdBN~\cite{2020Going} &  SNN training & ResNet34 & 6 & 63.72\%   \\ 
		\cline{2-6}
		& \multirow{2}{*}{SEW ResNet~\cite{2021Deep}} & \multirow{2}{*}{SNN training} & {ResNet18} & 4 & {63.18\%}   \\
		&  &  &											                             {ResNet34} & 4 & {67.04\%}   \\ 
		\cline{2-6}
		& \multirow{2}{*}{\textbf{Real Spike} } & \multirow{2}{*}{SNN training} & {ResNet18} & 4 & \textbf{63.68\%}$\pm 0.08$   \\
		&  &  &											                             {ResNet34} & 4 & \textbf{67.69\%}$\pm 0.07$   \\
		
		\hline			
		\multirow{4}{*}{CIFAR10-DVS}	
		& Rollout~\cite{2020Efficient} & Streaming & DenseNet & 10 & 66.80\%   \\	
		& STBP-tdBN~\cite{2020Going} & SNN training & ResNet19 & 10 & 67.80\%   \\  
		\cline{2-6}
		& \multirow{2}{*}{\textbf{Real Spike}} & \multirow{2}{*}{SNN training} & {ResNet19} 
		& 10 & \textbf{72.85\%}$\pm 0.12$   \\
		&  &  &											                                 {ResNet20} & 10 & \textbf{78.00\%}$\pm 0.10$   \\
		
		\bottomrule			   		         	            			         	
	\end{tabular}	
\end{table}

In this section, we compared the \textbf{Real Spike}-based SNNs with other state-of-the-art SNNs on several static datasets and a neuromorphic dataset. For each run, we report the mean accuracy as well as the standard deviation with 3 trials. Experimental results are shown in Tab.~\ref{tab:Comparison}.

\textbf{CIFAR-10(100).}
On CIFAR-10, our SNNs achieve higher accuracy than the other state-of-the-art methods, and the best result of 95.71\% top-1 accuracy is achieved by ResNet19 with 6 timesteps. And even trained with much fewer timesteps, \ie, 2, our ResNet19 still outperforms the STBP-tdBN under 6 timesteps with 0.85\% higher accuracy. This comparison shows that our method also enjoys the advantage of latency reduction for SNNs. On CIFAR-100, Real Spike also performs better and achieves a 2.53\% increment on ResNet20 and a 0.95\% increment on VGG16.

\textbf{ImageNet.}
The ResNet18 and ResNet34 were selected as the backbones. Considering that the image size of the samples is much larger, the channel-wise Real Spike was used. For a fair comparison, we made our architectures consistent with SEW-ResNets, which are not typical SNNs, where the IF model and modified residual structure are adopted. Results show that, even with 120 fewer epochs of training (200 for ours, 320 for SEW-ResNets), the channel-wise Real Spike-based SNNs can still outperform SEW-ResNets. In particular, our method achieves a 0.5\% increment on ResNet18 and a 0.65\% increment on ResNet34. Moreover, Real Spike-based ResNet34 with 4 timesteps still outperforms STBP-tdBN-based RersNet34 with 6 timesteps by 3.97\% higher accuracy.

\textbf{CIFAR10-DVS.}
To further verify the generalization of the Real Spike, we also conducted experiments on the neuromorphic dataset CIFAR10-DVS. Using ResNet20 as the backbone, Real Spike achieves the best performance with 78.00\% accuracy in 10 timesteps. For ResNet19, Real Spike obtains 5.05\% improvement compared with STBP-tdBN. It's worth noting that, as a more complex model, ResNet19 performs worse than ResNet20. This is because that neuromorphic datasets usually suffer much more noise than static ones, thus more complex models are easier to overfit on these noisier datasets.

\section{Conclusions}

In this work, we focused on the difference between SNNs and DNNs and speculated that the unshared convolution kernel-based SNNs would enjoy more advantages than those with shared convolution kernels. Motivated by this idea, we proposed \textbf{Real Spike}, which aims at enhancing the representation capacity for an SNN by learning real-valued spikes during training and transferring the rich representation capacity into inference-time SNN by re-parameterizing the shared convolution kernel to different ones. Furthermore, a series of \textbf{Real Spike}s in different granularities were explored, which are enjoy shared convolution kernels in both training and inference phases and friendly to both neuromorphic and non-neuromorphic hardware platforms. Proof of why \textbf{Real Spike} has a better performance than vanilla SNNs was provided. Extensive experiments verified that our proposed method consistently achieves better performance than the existed state-of-the-art SNNs.

\clearpage
%
%
\bibliographystyle{splncs04}
\bibliography{egbib}
\end{document}